# Comparison of Kinematics and Kinetics Between OpenCap and a Marker-Based Motion Capture System in Cycling

Reza Kakavand[1,2,3], Reza Ahmadi[1], Atousa Parsaei[1], W. Brent Edwards[1,2,3], Amin Komeili[1,2,3*]

[1]Department of Biomedical Engineering, Schulich School of Engineering, University of Calgary

[2]McCaig Institute for Bone and Joint Health, University of Calgary, Calgary, Canada.

[3]Human Performance Laboratory, Faculty of Kinesiology, University of Calgary, Calgary, Canada

**Corresponding author: Amin Komeili**

***** amin.komeili@ucalgary.ca

**Address: CCIT216, 2500 University Drive NW, Calgary, AB, T2N 1N4**

**Abstract**

This study evaluates the agreement of marker-based and markerless (OpenCap) motion capture systems in assessing joint kinematics and kinetics during cycling. Markerless systems, such as OpenCap, offer the advantage of capturing natural movements without physical markers, making them more practical for real-world applications. However, the agreement of OpenCap with a marker-based system, particularly in cycling, remains underexplored. Ten participants cycled at varying speeds and resistances while motion data were recorded using both systems. Key metrics, including joint angles, moments, and joint reaction loads, were computed using OpenSim and compared using root mean squared error (RMSE) per trial across participants, Pearson correlation coefficients (r) per trial across participants and repeated measures Bland-Altman to control trials' dependency within subject. Results revealed very strong agreement ($r > 0.9$) for hip (flexion/extension), knee (flexion/extension), and ankle (dorsiflexion/plantarflexion) joint angles. Relatively high RMSE values of 10.7° (±3.0°) and 12.4° (±4.6°) were observed for left and right ankle dorsiflexion/plantarflexion angles, respectively, per trial across participants. Knee flexion/extension RMSE per trial across participants were 9.3° (±3.8°) and 10.2° (±4.3°) for the left and right limbs, respectively. RMSE values of hip flexion/extension, adduction/abduction and rotation were 7.9° (±2.6°), 3.2° (±1.1°) and 3° (±1.2°) for the left side. Joint reaction forces and moments exhibited moderate to very strong agreement across most degrees of freedom. The RMSE of joint reaction forces ranged from 13.7 to 37.7 %BW. The hip medial-lateral moment had a minimum RMSE of 0.19 %BW×ht, while the highest RMSE of 1.27 %BW×ht was in the knee anterior-posterior moment. Despite strong overall agreement between the systems, variability per trial across participants in RMSE suggested that OpenCap may require further refinement in specific areas. These findings highlight the potential of markerless motion capture systems, such as OpenCap, in biomechanical analyses of cycling while also identifying areas where further refinement may be needed.

## 1 Introduction

Traditional marker-based systems require a controlled laboratory environment, specialized cameras, and relatively time-consuming experimental setup, limiting their utility in real-world or clinical settings [1–3]. Markerless motion capture presents several advantages over traditional marker-based motion capture systems. One key advantage of markerless systems is eliminating the need for skin marker placement, which can often lead to misplacement and skin motion artifacts. By removing the requirement for markers, markerless systems reduce issues such as noise filtering caused by missing markers and motion artifacts, thereby streamlining the data collection and analysis process [4–6]. Leveraging computer vision and deep learning algorithms, markerless motion capture directly detects human body landmarks from digital images [7–12]. For instance, Theia3D (Theia Markerless Inc., Kingston, ON, Canada) performs 3D pose estimation using 2D video data from standard video cameras, allowing data collection in diverse environments without the need for physical marker attachment [7]. This reduces manual processing steps and improves reproducibility in capturing 3D kinematics of more natural movements in real-world scenarios [7,13,14]. Moreover, markerless systems, such as DeepLabCut and Theia3D, are flexible and adaptable, capable of tracking various user-defined feature of interest, making them more accessible and versatile for various applications [12,15,16]. Markerless motion capture systems have the potential to facilitate large-scale studies and clinical applications, providing a scalable and economical solution for assessing human movement dynamics in diverse populations and environments [9]. For example, a recent study used data from a dataset of 628 individual Parkinson's disease patients across multiple clinical sites to develop a markerless motion capture system with smartphone and tablet devices. This system reliably assessed bradykinesia severity, closely aligning with clinician ratings and demonstrating its feasibility for integration into clinical workflows with minimal additional resources.

Recent advancements in human pose estimation algorithms, particularly open-source 2D pose estimation tools, such as OpenPose [17] and HRNet [18–21], have paved the way for the research community to implement markerless methods in data collection and measure kinematics [22–27]. While these machine learning models are application-specific and may lack generalizability [28], an alternative approach involves triangulating body key points identified by pose estimation algorithms across multiple videos and tracking these 3D positions with musculoskeletal models and physics-based simulations [29–34]. However, a significant limitation of this approach is the quality of the training data used for the 2D pose estimation, which can impact the expressiveness (how well the points reflect the actual movement) of the resulting 3D key points [35,36]. Commercial markerless motion capture systems often demand multiple wired cameras, proprietary software, and substantial computing resources, particularly for the analysis of long-duration data collection [9,37–39].

OpenCap[37] is an open-source, web-based software designed to provide free access to estimating the 3D kinematics and dynamics of human movement [37]. By integrating advancements in computer vision and musculoskeletal simulation, OpenCap enables the analysis of movement dynamics using videos captured by two or more smartphone cameras without the need for specialized hardware, software, or expertise. OpenCap utilizes smartphones, which are widely accessible, enabling motion capture outside of traditional lab settings. This makes biomechanical analysis more convenient and cost-effective compared to marker-based systems that require specialized equipment. With its cloud-based processing and ease of use, OpenCap allows for motion tracking in real-world environments, such as cycling, without complex setup. Additionally, OpenCap provides 3D pose estimations, which are preferred over the 2D pose capture methods used by Ceriola et al [40]. OpenCap mean absolute error of knee joint angle was reported in the range of 4.1° (2.3°–6.6°) during normal walking. However, the fidelity of OpenCap in quantifying 3D joint kinematics and dynamics for a pedaling task has not been fully verified against standard marker-based motion capture systems. Understanding the biomechanics of cycling, through the measurement of kinematics and kinetics, is crucial for optimizing pedaling performance, preventing injuries, and designing better equipment [41–43].

While markerless motion capture holds potential opportunities, it is essential to scrutinize differences in kinematic and kinetic measurements compared to the marker-based systems for studied activity. Assuming consistent measurements for different activities is not a legitimate approach. This is because activity type, equipment, environmental factors, and range and pace of the movement may affect the motion capture. This study aims to provide a comprehensive comparison of joint kinematics and dynamics in pedaling tasks using marker-based and markerless motion capture (OpenCap) systems, emphasizing the advantages and limitations of markerless technology in the biomechanical analysis of cycling.

## 2 Methods

### 2.1 Participants

Ten (5M/5F) healthy adult participants with an average age of 29.5 years (±3.3 SD), and a typical height of 1.76 m (±0.08 SD) and a body mass of 70.6 kg (±11.8 SD), were recruited for this study. Experiments were conducted in the Human Performance Lab at the University of Calgary. Subjects provided written informed consent, and the study received approval from the institutional ethics committee at the University of Calgary (REB #23-1803). Participants with any neuromuscular or musculoskeletal issues that might hinder cycling ability were excluded. Participants were instructed to wear tight, minimal clothing, and provided with cycling cleats (Santic, S3-KMS20025) to minimize interference during the tests.

## 2.2 Experimental setup and procedure

**Test protocol:** Subjects were asked to cycle at 90 ± 5.0 rpm (high velocity), and 60 ± 5.0 rpm (low velocity) and resistance levels of low, normal, and high, for two minutes each, followed by 5 minutes of rest between trials (participants were allowed to cycle at these velocities and resistances beforehand to familiarize themselves with the protocol). This resulted in 6 cycling powers ranging from 55 to 352 W at cyclists' preferred saddle height. The body movements were recorded by a Vicon (Vicon Motion Systems Ltd., Oxford, UK) and OpenCap system simultaneously.

***Marker-based*** motion capture included thirty-two reflective markers bilaterally affixed to the 2nd and 5th metatarsal heads, calcanei, medial and lateral malleoli, shank, medial and lateral femoral epicondyles, thigh, anterior and posterior superior iliac spines, and greater trochanters (Figure 1). These markers were tracked using a 10-camera motion capture system (Vicon Motion Systems Ltd., Oxford, UK) operating at a sampling rate of 250 Hz. A right-handed global reference system was defined with the positive x-axis in the anterior-posterior direction, a positive z-axis in the lateral-medial direction of the right limb of the participant, and positive y-axis in the vertical direction (Figure 1-B). A static calibration trial was collected for the marker-based motion capture data and used to scale the human body model. The trajectories of the markers were labeled to ensure they continuously tracked the correct positions of the lower body segments. Marker trajectory gaps of <0.03s were filled using cubic splines and pattern fill methods.

***OpenCap*** [37] was used to record video from four smartphones (iPhone 12 Pro, Apple Inc., Cupertino, CA, USA), with the HRNet pose detection algorithm operating at a sampling rate of 60 Hz. The smartphones were positioned 1.5 m above the ground, 3 m from the participant, arranged at 40° intervals around them (Figure 1). A 210×175 mm checkerboard was positioned within the view of all cameras to assist in computing the extrinsic parameters, including each camera's rotation and translation, during OpenCap's calibration step. Using a single image of the checkerboard, OpenCap automatically calculated these parameters to determine each camera's transformation relative to the global reference frame. Following calibration, OpenCap captured the participant in a stationary standing pose. Utilizing the anatomical marker positions inferred from static stance trial, OpenCap employed OpenSim's Scale tool to adjust a musculoskeletal model [44,45] to fit the participant's specific anthropometric measurements. The musculoskeletal model had 33 degrees of freedom, as described in the Figure 1. We conducted experiments using two to seven smartphones for data acquisition. Configurations with fewer than four smartphones resulted in unrealistic tracking body segments, primarily due to occlusions caused by the stationary bike's frame and handlebars. With four smartphones, the inverse kinematics results became more stable, and the motion capture appeared visually smooth. Configurations using five to seven smartphones deviated from OpenCap's recommended guidelines for optimal camera placement, which caution that camera angles positioned too closely can lead to unrealistic kinematic estimates. OpenCap advises arranging cameras symmetrically around the subject's movement direction, with a relative angle of 40–90° between cameras and at least 30° of separation from the sagittal plane. Based on these recommendations, we used four iPhone 12 (Apple Inc.) smartphones configured as illustrated in Figure 1.

***The model's constraints***: The default OpenCap musculoskeletal model [37] was used, where the femur articulates with the pelvis via a ball-and-socket joint. The orientation of the right femur relative to the pelvis was described using femur-fixed ZXY rotations, corresponding to hip flexion, adduction, and internal/external rotation, respectively. The same convention was applied to the left hip. The hip joint range of motion included 30° of extension to 120° of flexion, 50° of abduction to 30° of adduction, and 40° of external rotation to 40° of internal rotation.

The knee joint was modeled as a single degree-of-freedom hinge joint. The orientation of the right tibia relative to the right femur was determined by rotation about the femur-fixed –Z-axis, defined by the knee flexion angle, the same applied to the left knee. The coupled rotation and translation of the tibia with respect to the femur was parameterized by the knee flexion angle using joint kinematics defined by Walker et al. [46], modified such that the joint reference frames are aligned with anatomical approximations of the center of rotation. Similarly, patellar kinematics were defined as a function of the knee flexion angle using the model by Arnold et al. [47], modified such that the patella articulates with the femur. The knee flexion range spans from 0° to 120°.

The ankle, subtalar, and metatarsophalangeal joints were modeled as pin joints, representing ankle dorsiflexion, subtalar inversion, and toe flexion, respectively. Joint axes were aligned with anatomical measurements reported by Inman et al. [48]. The ranges of motion were 40° of plantarflexion to 30° of dorsiflexion, 20° of eversion to 20° of inversion, and 30° of toe extension to 30° of toe flexion.

**In the scaling step**, we tried to customize a generic model to match an individual's anthropometric dimensions by adjusting segment lengths and inertial properties based on subject-specific data, such as marker trajectories from a static trial and the subject's mass [49]. This process ensures that the model's geometry and mass distribution closely reflect the actual subject size. The total squared error, RMSE, and maximum error for the scaling step were 0.06 (±0.05), 0.04 (±0.02) and 0.09 (±0.03) m, respectively. The **Inverse Kinematics** total squared error, RMSE, and maximum error were 0.015 (±0.01), 0.02 (±0.007) and 0.04 (±0.018) m, respectively.

***Pedal reaction force*** and crank position were measured at a sampling rate of 250 Hz using Sensix pedals (I-Crankset system, Ver. 4.8.5, SENSIX, Poitiers, France at https://sensix.fr/pedal-sensors) and an encoder (LM13ICD20DA10A00, RLS), respectively. Pedal force data were low-pass filtered using a fourth-order, zero-lag Butterworth filter at 3 Hz [50]. To address the inability to directly synchronize the OpenCap system with the pedal force data acquisition system, the top dead center (TDC) position of the crank was used as a common temporal landmark. In the pedal force system, TDC was measured using an encoder, while in OpenCap, it was visually identified based on the 5th metatarsal marker reaching its highest position (since the cycling shoes were secured to the pedals directly beneath the metatarsals). This approach allowed alignment of the data between the two systems. To ensure temporal correspondence for analysis, the OpenCap data were subsequently interpolated to a sampling frequency of 720 Hz, effectively providing two data points per degree of crank revolution.

We utilized an identical modeling and simulation pipeline for both marker-based and OpenCap data. *Kinematic and kinetic* data were calculated from markers' position and pedal force data in OpenSim 4.4 software. OpenSim's Scale and Inverse Kinematics tools were used to adjust the musculoskeletal models and to calculate joint kinematics, respectively. Subsequently, joint kinetics were computed from joint kinematics (filtered at the same frequencies as the pedal force data) and pedal force data using the Inverse Dynamics tool and Joint Reaction analyses in OpenSim.

### 2.3 Statistical Analysis

The root mean squared error (RMSE) and the Pearson correlation coefficients (r) [51] with 95% CI were computed between the marker-based motion capture and OpenCap [52] for hip (flexion/extension, adduction/abduction, and rotation), knee (flexion/extension) and ankle (dorsiflexion/plantarflexion and supination/pronation) joint angles, joint moment (net moment acting on the joint), reaction force and reaction moment (the internal joint load resulted from articular surfaces contact and ligament). Anatomical angles were measured with respect to the standing posture. We denoted r-values in the range of 0 to 0.36 as poor, 0.36 to 0.67 as moderate, 0.67 to 0.9 as strong, and 0.9 to 1.0 as very strong agreement [53]. RMSE and r-values were computed for each trial to control data dependency, and a boxplot was created to evaluate their median, mean, and variability. Repeated measures Bland-Altman analysis (using a fixed-effects ANOVA method to account for the dependency of trials within each subject) with 95% limits of agreement was employed to compare the outcome of two methods in one revolution [54,55]. A power analysis was conducted using G*Power 3.1 to estimate the minimum required sample size for comparing joint kinematics between OpenCap and the gold standard marker-based system [56,57]. Given the within-subject design of the study, where each participant's kinematics was measured using both systems across 6 cycling power output conditions, a paired-samples t-test was selected. To estimate effect sizes, we used data from previous studies reporting marker-based measurements. Specifically, we referenced reported values of 72.6 ± 2.2° for hip flexion, 110.8 ± 2.2° for knee flexion, and 18.4 ± 4.0° for ankle dorsiflexion [58]. However, recognizing that effect sizes could vary across power output conditions, we calculated the effect size for each joint angle within each of the 6 power output conditions. The smallest observed effect size across the 6 conditions for each joint was then used for the power analysis. This ensured that the sample size was sufficient to detect even the smallest anticipated differences. For the ankle dorsiflexion, the smallest observed effect size across the six power output conditions was 1.075. For the hip flexion, the smallest effect size was 1.36, and for knee flexion, it was 1.33. Using a two-tailed test with an alpha level of 0.05 and a statistical power of 0.80, the power analysis indicated that the maximum required sample size was 9 participants, which was necessary to detect differences in ankle dorsiflexion. To ensure adequate power across all joints and 6 power output conditions, a conservative sample size of 10 participants was selected for the study. The paired t-test assumes independence of observations between participants, which we consider to be met in our study design.

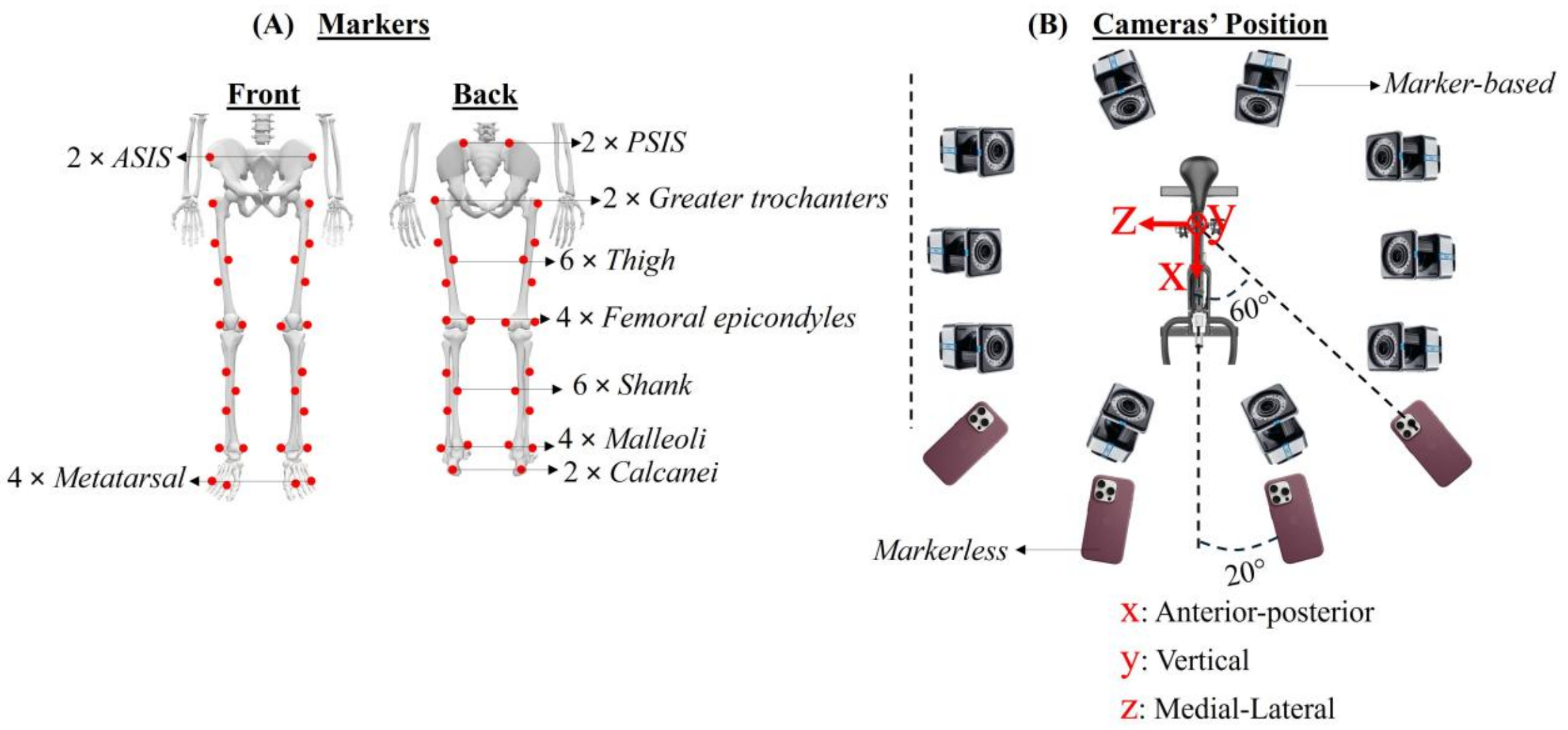

Figure 1. A) Marker placement of the lower extremity limbs, and B) position of cameras for OpenCap (4 smartphones) and for marker-based (Vicon cameras) motion capture systems.

## 3 Results

The vertical, anterior-posterior, and lateral-medial pedal reaction force components are illustrated in Figure 2, with the largest component being the vertical component at around the midpoint between Top Dead Center (TDC) and Bottom Dead Center (BDC) position. The smallest force component was the medial-lateral, which was an order of magnitude smaller than the vertical force component. The maximum hip and knee extension angles occurred around the Bottom Dead Center (BDC), where the distance between the hip and ankle joint is maximum during the pedaling task (Figure 3). Similarly, the maximum flexion angles of hip and knee joints occurred around TDC. Mean Hip Flex/Ext angles ranged from 44° (BDC) to 81° (TDC), and Mean knee Flex/Ext ranged from 47° (BDC) to 105° (TDC) for the left leg. Hip abduction angle was highest at BDC, coincident with maximum hip joint rotation angle. Hip Flex/Ext moment followed similar trends to the ankle Dors/Plnt moment. For example, hip extension moment reached its maximum at 83° and ankle plantarflexion at 100° of cycle for the left leg. The maximum knee extension moment occurred close to TDC, while the maximum knee flexion moment was at around 35 % of the left leg revolution at crank angle of 126° (or 83 % for the right leg) for both motion capture systems (Figure 3).

RMSE values for joint angles when predicted using the OpenCap and marker-based systems were 7.9° for hip flexion/extension and 3.2° for hip adduction/abduction and rotation for the left leg (Figure 4). Higher RMSE values (~10°) were observed for knee and ankle joint angles (Figure 4). Similar RMSE in angles were observed between right and left sides. For example, RMSE of 10.2° and 9.3° was obtained for the right and left knee Flex/Ext angles. Joint moments had RMSE values in the range of 0.44–0.89 %BW×ht for hip and knee moments, and 1.03–1.4 %BW×ht for ankle moments (Figure 4). Very strong agreement ($r > 0.9$) was noted for hip flexion/extension, knee flexion/extension, and ankle dorsiflexion/plantarflexion, while hip adduction/abduction and ankle supination/pronation showed strong agreement for both right and left limbs ($0.7 < r < 0.9$) (Figure 4). The Bland-Altman plot showed no clear trend in the differences across the range of angle and moment measurements, indicating no proportional bias (Figure 5). However, several data points fell outside the limits of agreement, highlighting instances where the two methods produced discrepant results. An overall underprediction of ankle angles by OpenCap compared to the marker-based method was observed, as indicated by positive differences (Figure 5). The difference in ankle moment prediction increased as the mean increased. While no consistent relationship was observed between RMSE and changes in joint angles and moments across all conditions, the RMSE for some joints increased with higher cadences and resistances (Figure 6). For instance, the RMSE for the knee flexion/extension moment increased from 0.6 %BW×ht in Trial 3 (60 rpm, normal resistance) to 2.4 %BW×ht in Trial 4 (90 rpm, normal resistance), and further to 3.6 %BW×ht in Trial 6 (90 rpm, high resistance).

Joint reaction force (%BW) and joint reaction moment (%BW×ht) were plotted in Figure 7. The vertical reaction force (Fy) of knee and ankle joints reached its maximum at 90° (75% shown in Figure 7) where the maximum force against the pedal was exerted (Figure 2) [59]. The vertical hip joint reaction force reached its absolute maximum at BDC, with values ranging from -60% to -140% BW. The mean anterior-posterior hip joint reaction force was approximately 20% BW at BDC and peaked at 76% BW at 65° (67% of the cycle), just before the push-down phase. (shown in Figure 7). The relatively lower medial-lateral joint reaction forces, i.e., in the z-direction, were expected since the minimum pedal force occurs in this direction. The RMSE of joint reaction forces ranged from 13.7 to 37.7 %BW (Figure 8). The highest RMSE of 37.7 %BW was observed in the mean knee joint vertical reaction force (Figure 8). Very strong agreement ($r > 0.9$) was found for hip, knee, and ankle vertical forces Fy, with strong to moderate agreement for other directions (Figure 8). For example, r-value of 0.95 was found for ankle joint vertical reaction force, while 0.44 and 0.51 r-values were obtained for its anterior-posterior and medial-lateral reaction forces, respectively (Figure 8). Hip flexion/extension moment Mz had a minimum RMSE of 0.19 %BW×ht, while the highest RMSE of 1.27 %BW×ht was in the knee adduction/abduction moment Mx (Figure 8). The Bland-Altman plot showed that, while the bias was small for joint reaction force (%BW) and moment (%BW×ht), the width of the limits of agreement indicated considerable variability in the differences between the two methods (Figure 9). A noticeable trend in the differences was observed, particularly in the medial-lateral direction.

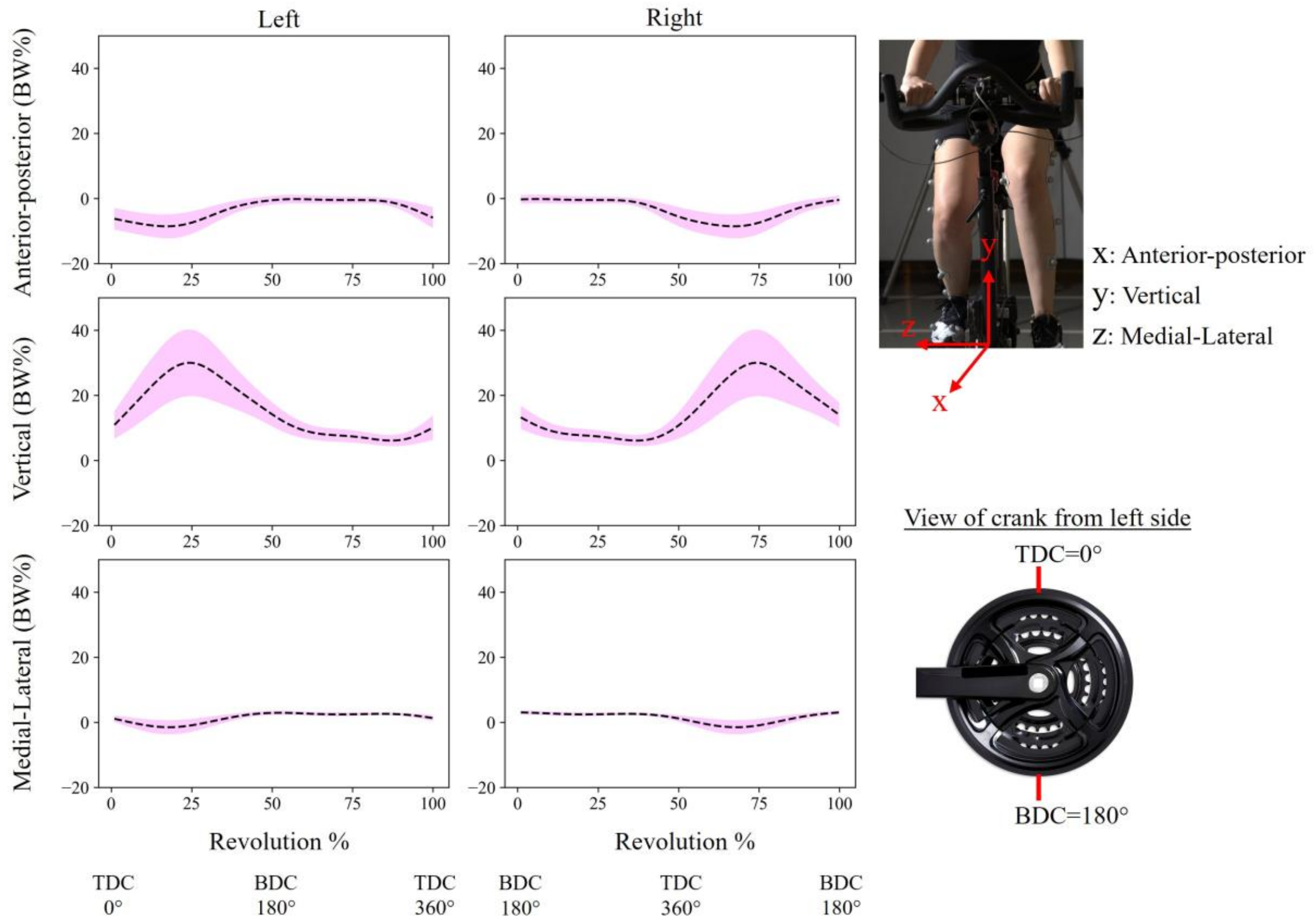

Figure 2. The mean and SD of pedal force components over all trials are plotted in anterior-posterior (x-direction), vertical (y-direction), and medial-lateral directions (z-direction) that are fixed with respect to ground. Left and right columns correspond to the left and right pedals. The cycling revolution for the left leg began at TDC, while for the right leg, it started at BDC.

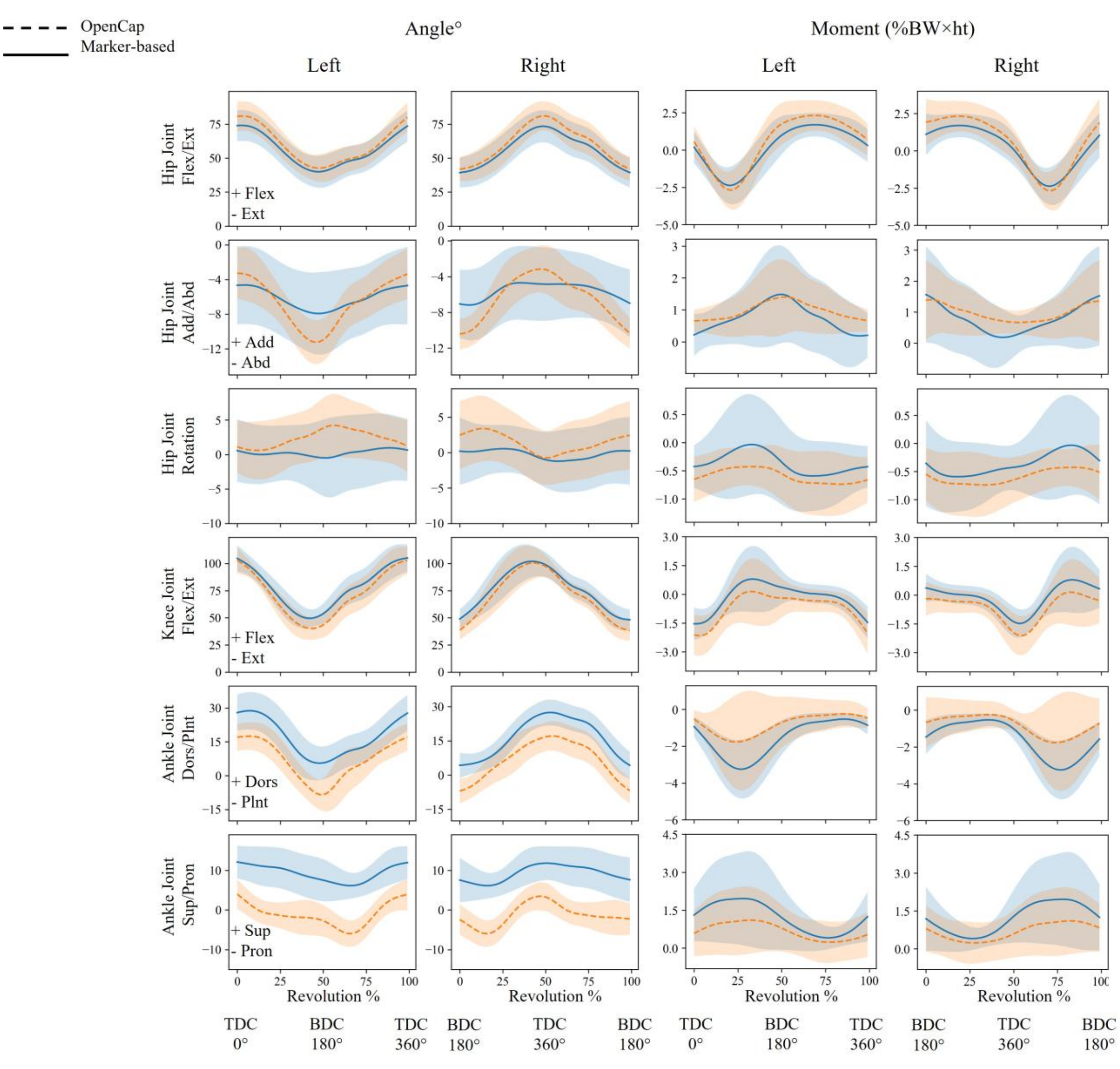

Figure 3. Joint angle (°) and moment (%BW×ht) computed from Inverse Kinematics and Inverse Dynamics analyses, respectively, in OpenSim. Anatomical angles were measured with respect to the standing posture.

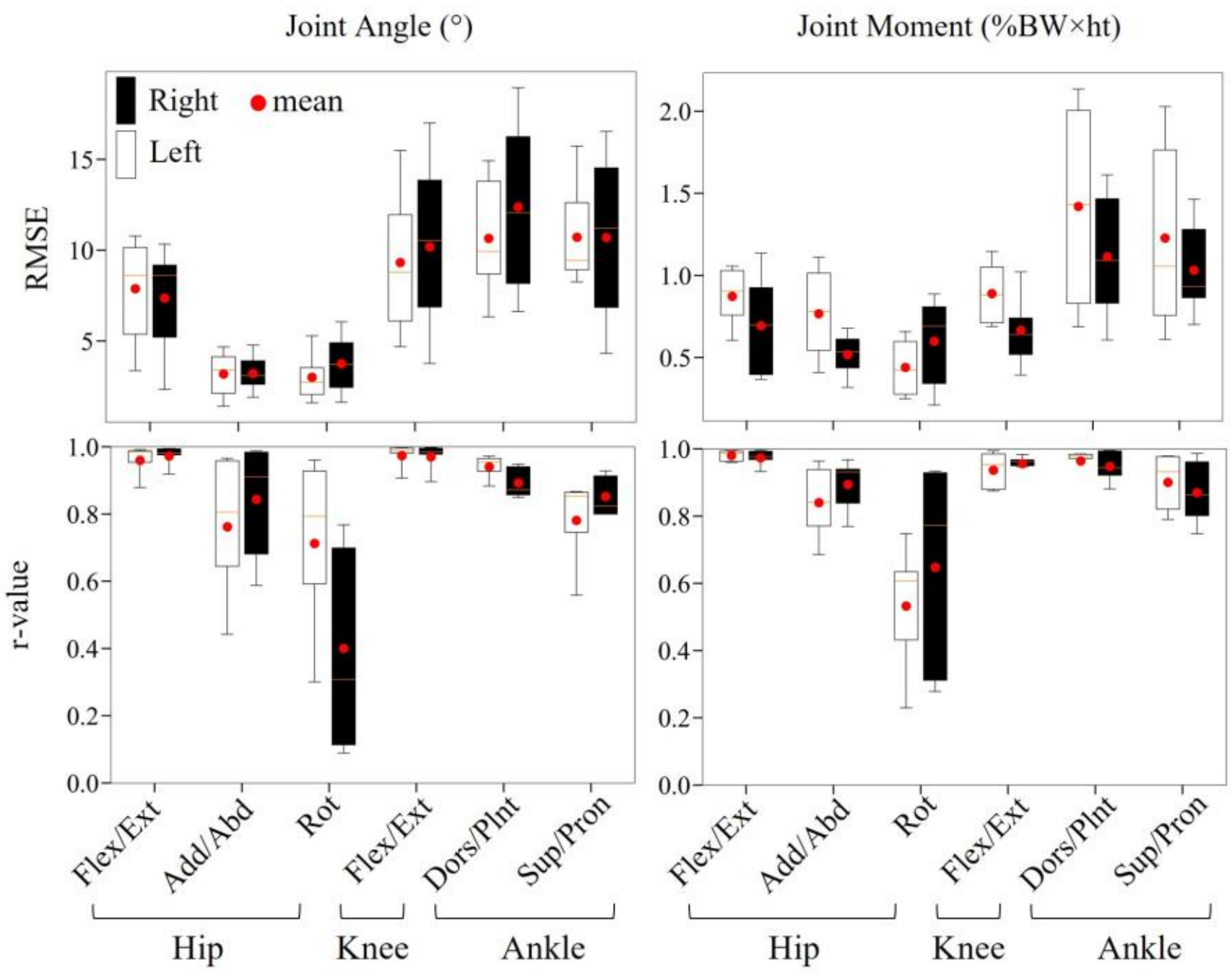

Figure 4. RMSE and Pearson correlation coefficient (r-value) with 95% CI for joint angle (°) and moment (%BW×ht) when computed from OpenCap and maker-based methods.

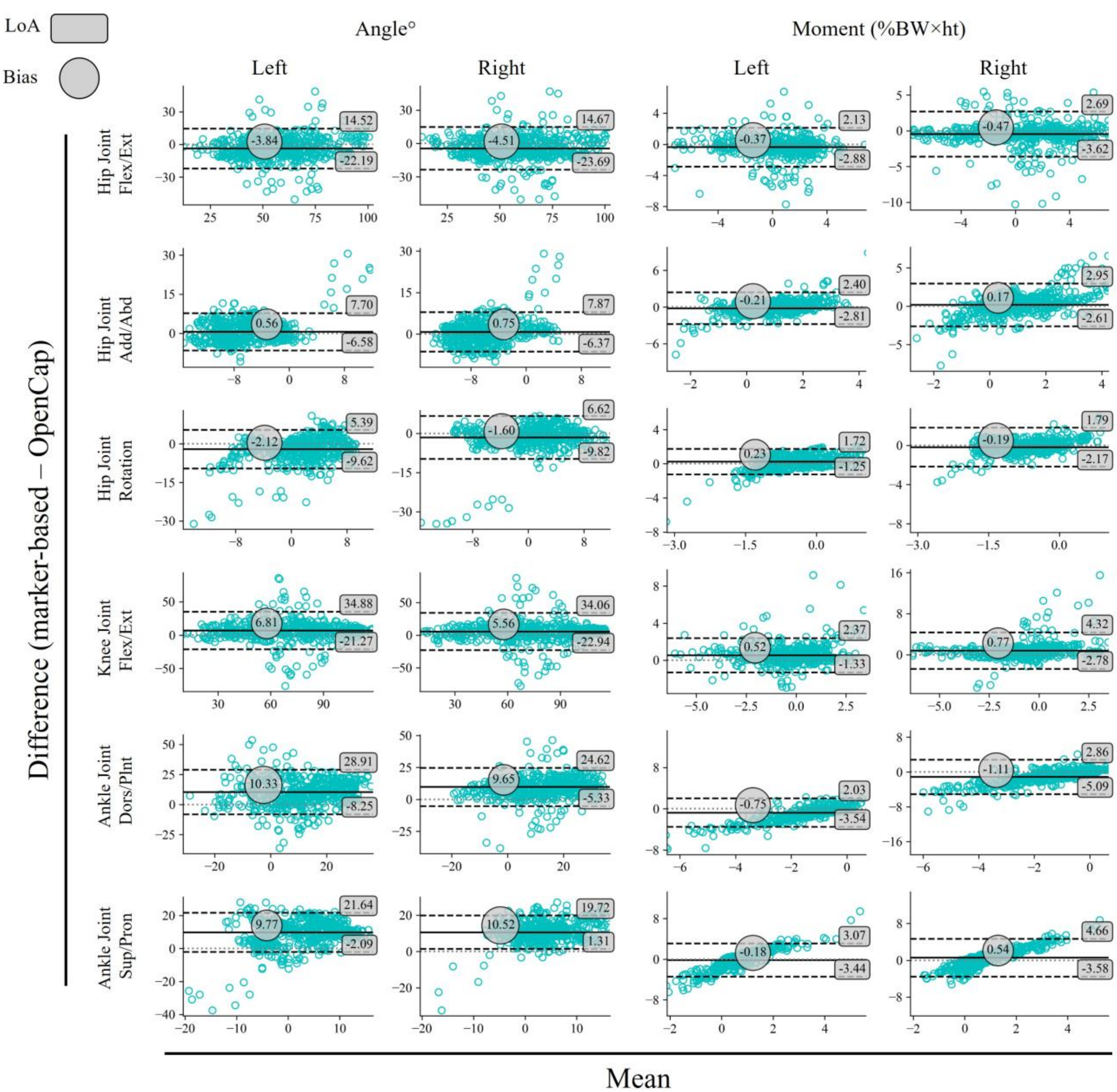

Figure 5. Repeated measures Bland-Altman plots between OpenCap and marker-based methods for joint angle (°) and moment (%BW×ht) for one crank revolution. Dashed lines indicate 95% Limits of Agreement (LoA), and the solid lines show the mean bias.

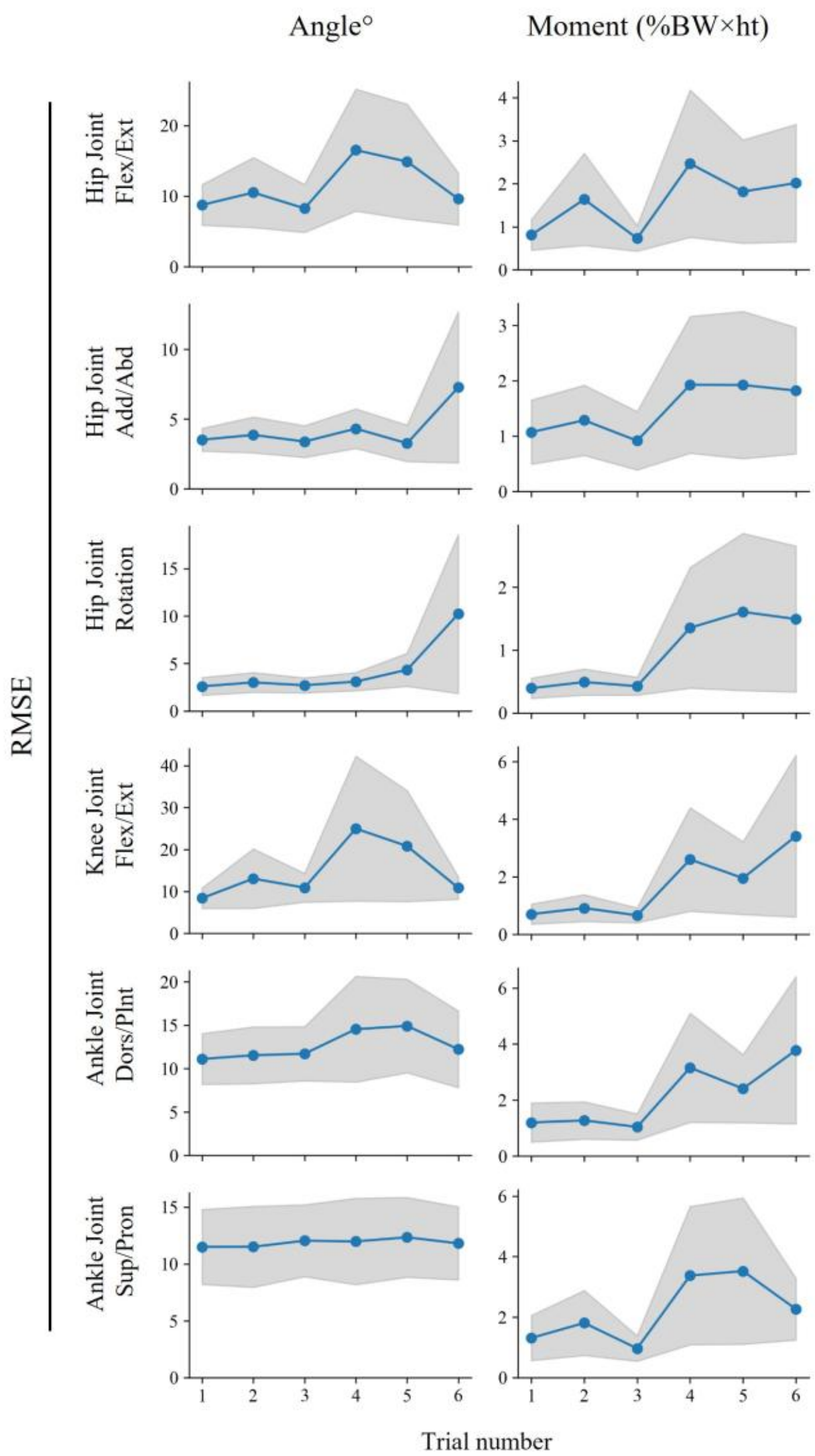

Figure 6. RMSE with 95% CI as a function of trial number for joint angle (°) and moment (%BW×ht) when computed from OpenCap and maker-based methods for the right limb. Trial 1: 60 ± 5.0 rpm with low resistance; trial 2: 90 ± 5.0 rpm with low resistance; trial 3: 60 ± 5.0 rpm with normal resistance; trial 4: 90 ± 5.0 rpm with normal resistance; trial 5: 60 ± 5.0 rpm with high resistance; trial 6: 90 ± 5.0 rpm with high resistance.

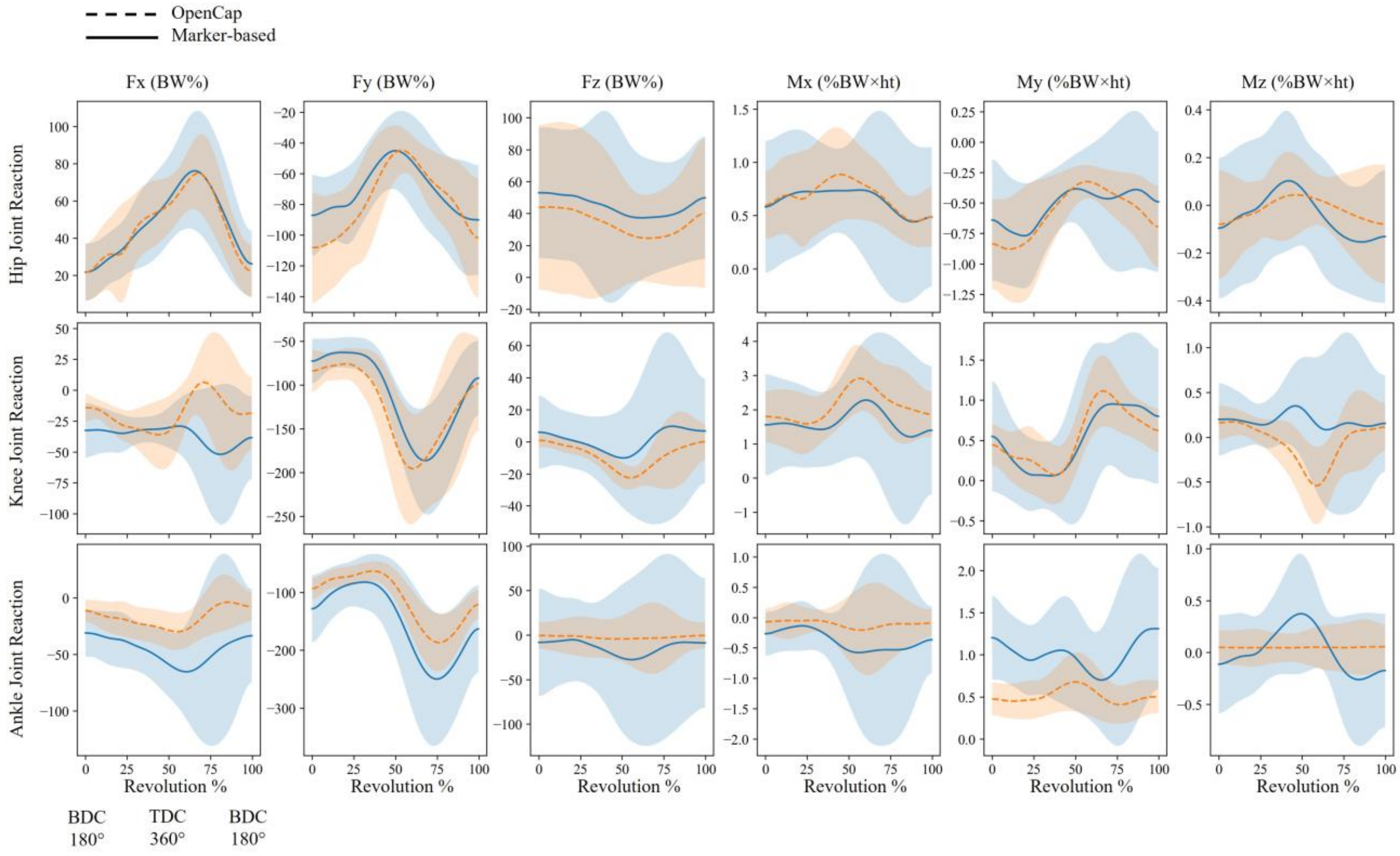

Figure 7. Joint reaction force (%BW) and joint reaction moment (%BW×ht) components computed from the kinematics of motion recorded by OpenCap and marker-based systems using Joint Reaction analysis in OpenSim. Results of the right leg were shown with the coordinate system defined based on Figure 1.

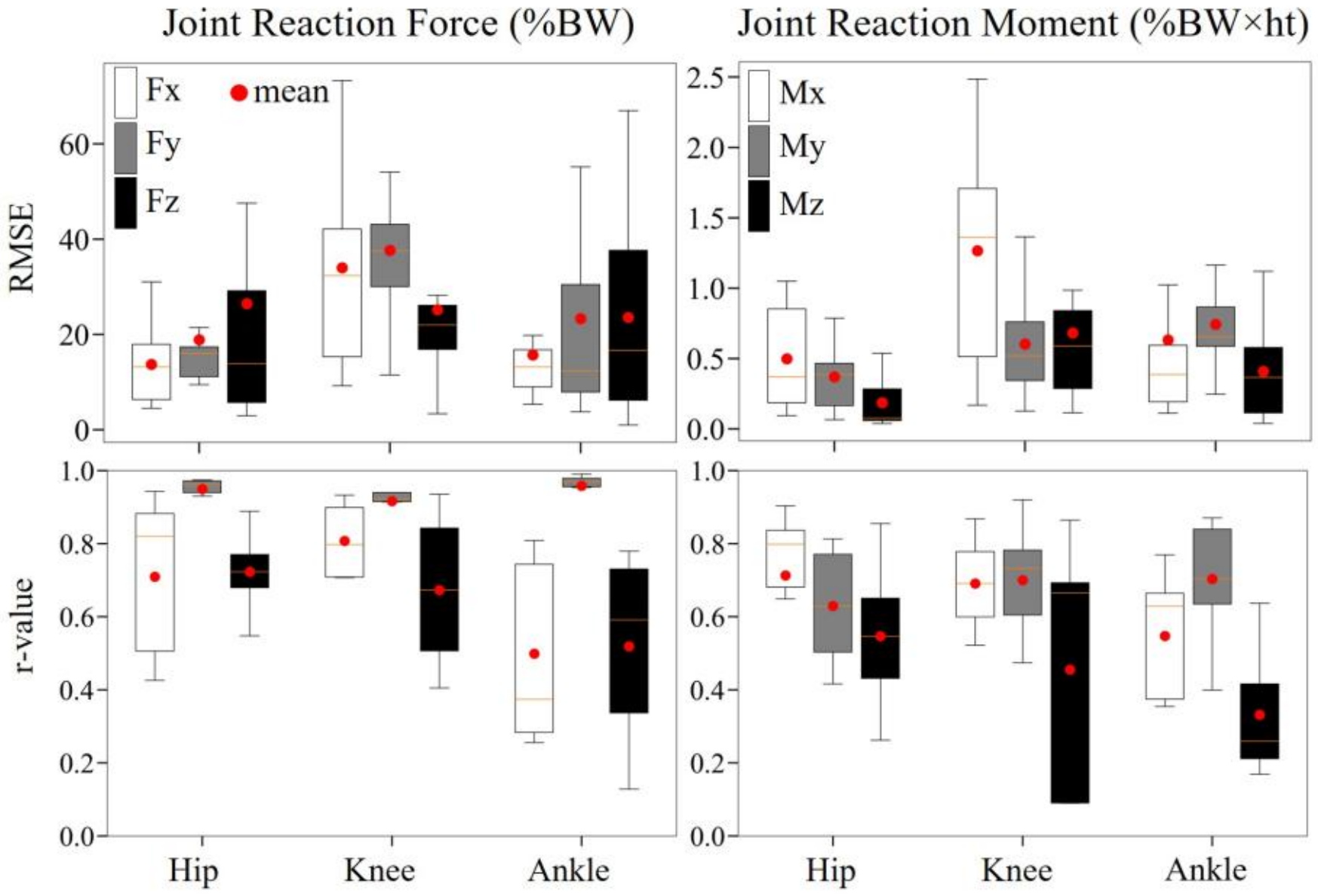

Figure 8. RMSE and Pearson correlation coefficient (r-value) with 95% CI for joint reaction force (%BW) and joint reaction moment (%BW×ht) computed from the kinematics of motion recorded by OpenCap and marker-based systems. The coordinate system was defined based on Figure 1.

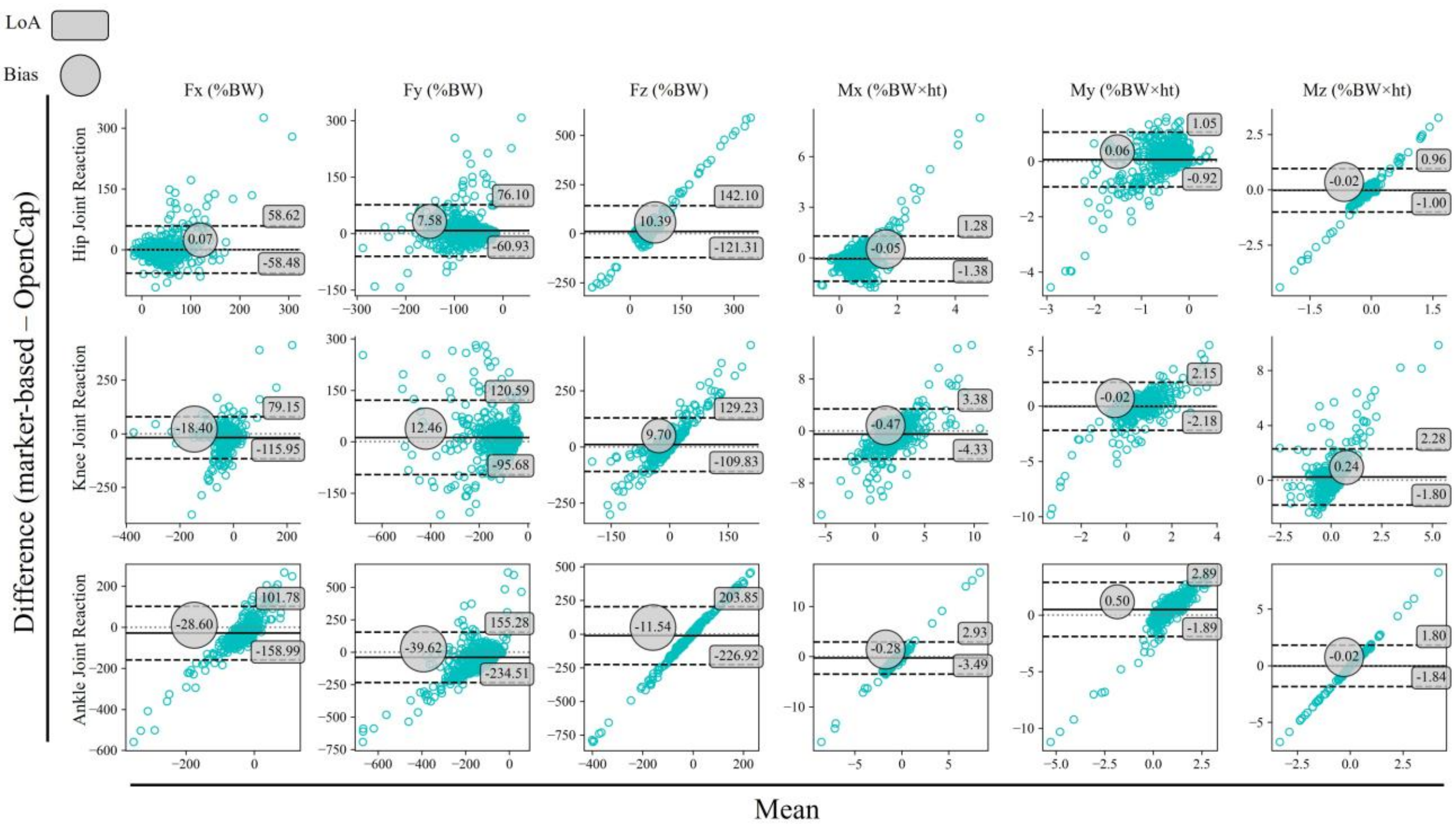

Figure 9. Repeated measures Bland-Altman plots between OpenCap and marker-based methods for joint reaction force (%BW) and joint reaction moment (%BW×ht). Dashed lines indicate 95% Limits of Agreement (LoA), and the solid lines show the mean bias.

## 4 Discussion:

The purpose of this study was to compare the joint kinematics and kinetics of the pedaling task measured by a traditional marker-based and a markerless motion capture system (OpenCap). While both systems provide valuable insights into human movement dynamics, our findings highlighted notable differences across key biomechanical metrics.

The comparison of joint angles between the two systems showed high agreement for most degrees of freedom, particularly for hip flexion/extension, knee flexion/extension, and ankle dorsiflexion/plantarflexion angles, where correlation values exceeded 0.9, as shown in Figure 3 and Figure 4. The very strong correlations observed for sagittal plane angles suggest that OpenCap provides estimates of joint angles during cycling that are closely aligned with those obtained from marker-based systems. However, the relatively high RMSE values and notable discrepancies in some joint angle estimates (Figure 5) indicate that there may still be room for improvement in achieving consistent agreement across all kinematic measures. For instance, measurements from the OpenCap and marker-based systems at BDC for joint angles with small variations, such as hip rotation and ankle supination/pronation (Figure 3), did not agree, although their variation trend matches. Similar levels of disagreement in local measurements between the two systems were observed for joint moments. For instance, larger discrepancies in hip rotation moments and ankle dorsiflexion/plantar flexion moments were noted when the crank was parallel to the ground at 90°. This corresponds to the middle of the power phase, where maximum power output occurs (Figure 2). Another notable point is that the spatial distribution of differences in the measurements of joint angles and moments does not follow the same trend (Figure 3). For instance, a large difference in the hip joint adduction/abduction angle was observed between the two systems at BDC, but its corresponding moment showed high agreement. This was expected, as the calculation of joint moments is affected not only by the corresponding joint angle but also by other joint angles, due to the interaction of moment arms in the moment equilibrium equations used in inverse dynamics. The variability of measuring kinetics and kinematics between the two systems might be attributed to the implemented neural network, which is an LSTM model in OpenCap, affecting its expressiveness (how well the points reflect the actual movement) in locating the trajectory of limb segments. This LSTM model was intended to translate synchronized 3D key points into a comprehensive anatomical marker set, leveraging temporal patterns in motion data for tracking and representation. The model might be improved by enhancing the diversity and volume of training examples.

The error between OpenCap and marker-based system observed in this study (RMSE < 7.9° for hip flexion/extension angle, < 10.2° for knee flexion/extension angle, and <12.4° for ankle dorsi/plantar flexion angle) is comparable to that of other markerless systems used in previous studies [50,60–63]. Serrancoli et al. [50] reported a RMSE < 3° for hip flexion/extension, <5° for knee flexion/extension, and <11.5° for dorsi/plantar flexion ankle joints during cycling using a custom-built markerless motion capture system. Among all the joints analyzed, Castelli et al. [64] recorded the highest error at the hip flexion/extension, with a RMSE of 6.1° . Ceseracciu et al. [65] reported RMSE values during gait, with 17.6° at the hip flexion/extension, 11.8° at the knee flexion/extension, and 7.2° at the ankle dorsi/plantar flexion using a custom-built markerless motion capture system. Additionally, Uhlrich et al.[37] obtained a mean absolute error of 1.7° to 10.3° for joint angles during walking, squat, sit-to-stand and drop jump between OpenCap and marker-based motion capture. Therefore, these findings suggest that the error associated with OpenCap in this study is comparable to the levels reported for other markerless systems in the literature.

Joint moments, which represent net moments acting on the joint, were computed using Inverse Dynamics (Figure 3). The results showed a high level of agreement (r > 0.9) between the two systems for joint moments in the sagittal plane. Lower r values for parameters in the coronal and transverse planes, such as hip Add/Abd moment, reflect a significant degree of variation between the two systems in the calculation of corresponding parameters. One possible reason for these discrepancies is the sensitivity of systems to small variations in joint angle trajectories. For instance, hip Add/Abd and rotation varied on average less than 6° during cycling.  OpenCap's reliance on computer vision and deep learning models to estimate 3D kinematics from 2D video might introduce small errors that accumulate for proximal joints during Inverse

Dynamics analysis. The propagation of these small kinematic errors through the inverse dynamics calculations, especially for degrees of freedom with smaller ranges of motion, likely explains the reduced agreement observed for hip adduction/abduction and rotation moments. A further limitation of markerless motion capture systems is the relatively sparse number of keypoints identified. Increasing the density of keypoints would likely improve the accuracy of describing movements, particularly in non-sagittal planes (e.g., frontal and transverse), where subtle motions may be missed with the current keypoint sparsity [66,67].

To assess the agreement between the OpenCap motion capture system with established literature, we examined the joint moment profiles at the hip, knee, and ankle, comparing our results to those reported in previous studies. The hip (flexion/extension) moment followed a similar pattern to other studies [62,68], with the maximum extensor moment occurring around 90° of the revolution (Figure 3). The knee moment has a large extension at TDC and increases during the first half of the cycle, reaching its peak around 126° [50,62] (Figure 3). The ankle dorsiflexion/plantarflexion moment from marker-based motion capture system remains in plantarflexion throughout the cycle, with its peak occurring during push down phase when crank is horizontal at 100°, i.e., 28% of cycle for left leg and 78% of cycle for right leg [50,62] (Figure 3).

The right and left joints exhibited similar RMSE and r values across most parameters, except for hip joint rotation, demonstrating the consistency of the markerless system in capturing repeated 3D motion for both legs. We propose that if the human body is approximated as a cylinder with its centerline aligned dorsoproximally, hip rotation would occur around this centerline. This type of rotation results in minimal displacement of the body's volume in recorded videos compared to rotations in the sagittal plane. Consequently, the joint detection and tracking algorithm (HRNet) in the OpenCap system may exhibit greater uncertainty in predicting body segment positions. Moreover, these limitations may be exacerbated by factors such as the relatively larger amount of soft tissue surrounding the hip and occlusions caused by the handlebar during dynamic activities. Despite overall strong agreement of OpenCap with the marker-based method in measuring the kinematics of the pedaling task in the sagittal plane, measurements of parameters in other planes had considerable discrepancy with respect to the marker-based method highlighting the challenges associated with capturing three-dimensional joint kinematics using markerless systems. This discrepancy may highlight the need for further refinement in algorithmic processing, camera placement, and calibration.

The Bland-Altman plots for joint moments and reaction forces revealed clear heteroscedasticity, which was particularly noticeable for ankle joint reaction forces and moments, hip and knee joint medial-lateral reaction forces, and hip and knee joint reaction moments (Figure 9). This indicates that the agreement between the marker-based and markerless (OpenCap) systems for these specific kinetic variables was not uniform across the range of measured values during cycling. Specifically, the variability in the differences between the two systems tended to increase with the magnitude of these forces and moments. This non-constant variance has important implications for the interpretation of LoA. The large LoA observed for these specific joints and degrees of freedom, coupled with the pronounced heteroscedasticity, suggests that the magnitude of the discrepancies between the two systems can be substantial, particularly at higher loads and velocities encountered during more demanding phases of the cycling task. Therefore, while OpenCap demonstrates potential for broader biomechanical analysis in cycling, the marked heteroscedasticity and large LoA for ankle joint reaction forces and moments, hip and knee medial-lateral reaction forces, and hip and knee reaction moments warrant careful consideration when interpreting results for these specific variables.

While markerless systems like OpenCap offer significant advantages in terms of practicality, accessibility, user experience, and reduced experimental setup, their agreement with traditional marker-based systems in capturing certain biomechanical parameters (Figure 5 and Figure 9) and also under more demanding cycling conditions (higher cadences and resistances as shown in Figure 6) still falls short. However, for large-scale applications where quick, efficient data collection and the estimation of the changes in joint biomechanical parameters are crucial, the trade-off between accessibility and precision might be justified. For example, OpenCap successfully distinguished the differences in movement dynamics between young and older adults during rising from a chair [37], despite its limited agreement with marker-based systems in measuring absolute value of kinematics. With the emergence of advanced image processing, artificial intelligence methods, and more advanced portable cameras, the agreement between markerless systems—such as OpenCap—and traditional motion capture methods continues to improve, making them a viable solution for easier and more accessible motion analysis in the field of human biomechanics.

While OpenCap shows agreement with marker-based systems for sagittal plane kinematics and kinetics, our findings point to specific areas where further refinement of the underlying algorithms may be necessary. Strengthening the robustness of the 2D keypoint detection (HRNet) in handling occlusions and capturing subtle movements, alongside improvements in the 3D pose estimation (LSTM model) to better reconstruct out-of-plane motion and track proximal joints, could make significant improvements. For instance, incorporating attention mechanisms, as explored by Zhang et al.[69], could allow HRNet to focus on pertinent image regions and better handle occlusions, potentially leading to more robust keypoint detection. Considering the sensitivity of inverse dynamics to small kinematic variations, particularly for degrees of freedom with limited ranges of motion, even incremental enhancements in the precision of joint angle estimates could lead to more consistent moment calculations [9,70]. Addressing these aspects has the potential to narrow the observed differences, especially for coronal and transverse plane parameters, and further solidify the utility of markerless systems like OpenCap in biomechanical analysis.

However, marker-based systems, often considered the gold standard, are also subjected to errors[71–73]. These can include errors from marker misplacement, inconsistency in marker placements between individuals, movement of the skin relative to the underlying bone. Therefore, the discrepancies observed between OpenCap and the marker-based system in this study likely reflect a combination of errors from both methodologies.

One limitation of our study is the restricted number of smartphones we could utilize in OpenCap setup. Given the specific positioning requirements and distance limitations between each smartphone, we determined that four smartphones are the optimal number to ensure consistent and stable data collection. Additionally, the placement of smartphones was restricted to within +/- 60° from the anterior position of the participant (Figure 1), which prevented us from positioning the smartphones around the rider as is typically done with marker-based motion capture systems. This limitation may have affected the quality of data collection and potentially limited the consistency of our biomechanical analyses. Additionally, when using OpenCap system, one notable difficulty raised when dealing with occlusions caused by the bicycle frame [74,75]. This may also be a source of error observed between OpenCap and marker-based systems measurements. We recognize that the limited sample size of ten healthy adults (5 males, 5 females) with specific average age, height, and body mass characteristics restricts the generalizability of our conclusions. Furthermore, the experimental design, while allowing for a controlled comparison between OpenCap and marker-based motion capture, employed a specific cycling protocol with set cadences (60 ± 5.0 rpm and 90 ± 5.0 rpm) and resistance levels, resulting in cycling powers ranging from 55 to 352 W. This controlled protocol, while necessary for experimental rigor, may not fully capture the range of cadences, resistances, and power outputs encountered in diverse cycling activities (e.g., sprinting, hill climbing, long-distance riding). Consequently, the findings might be most applicable to cycling performed under similar, controlled laboratory conditions. Future research should prioritize larger and more diverse samples, encompassing a broader range of participant characteristics and cycling protocols, including variations in cadence, resistance, and power output, as well as data collection in field settings, to enhance the generalizability of findings related to markerless motion capture in cycling biomechanics.

In conclusion, while OpenCap offers significant advantages in practicality, accessibility, and ease of use, its current disagreement with marker-based motion capture system in measuring certain joint angles, such as hip adduction/abduction, rotation, and ankle supination/pronation, necessitate caution in applications requiring high precision. Specifically, the error margins observed in this study may limit the immediate applicability of OpenCap for precise quantitative assessments. Nonetheless, for large-scale biomechanics studies where the focus may be on capturing general movement patterns and trends across populations, OpenCap would be a valuable tool. In such contexts, its minimal equipment dependency and ease of use can be advantageous.

**Acknowledgment:** The project was supported by the Natural Sciences and Engineering Research Council Canada (NSERC) Discovery grant [grant number 401610]

## Author contributions:

R.K.: conceptualization, data acquisition, motion analysis, writing, visualization, statistical analysis, project management; R.A.: data acquisition, motion analysis; A.P.: data acquisition, motion analysis; W.B.E.: management and writing; A.K.: conceptualization, funding, management, writing, visualization, motion analysis, project administration.

**Conflict of interest:** None